\documentclass[lettersize,journal]{IEEEtran}
\usepackage{amsmath,amsfonts}
\usepackage{algorithmic}
\usepackage{algorithm}
\usepackage{array}
\usepackage[caption=false,font=normalsize,labelfont=sf,textfont=sf]{subfig}
\usepackage{textcomp}
\usepackage{stfloats}
\usepackage{url}
\usepackage{verbatim}
\usepackage{graphicx}
\usepackage{cite}
\usepackage{amsmath}
\usepackage{comment}
\usepackage{wrapfig}
\usepackage{diagbox}
\usepackage{booktabs}
\usepackage{siunitx}
\usepackage{colortbl}
\usepackage{xcolor}
\usepackage{enumitem}
\usepackage{arydshln} 
\usepackage{multirow}
\usepackage{tabularx}
\usepackage{orcidlink}
\usepackage{balance}
\begin{document}
\title{Pseudo Triplet Guided Few-shot Composed Image Retrieval}

\author{Bohan Hou\orcidlink{0009-0008-0751-166X}, Haoqiang Lin\orcidlink{0009-0000-5768-5467}, Haokun Wen\orcidlink{0000-0003-0633-3722}, Meng Liu\orcidlink{0000-0002-1582-5764}, Mingzhu Xu\orcidlink{0000-0002-1492-0970} and Xuemeng Song\orcidlink{0000-0002-5274-4197}
\thanks{Bohan Hou, Haoqiang Lin and Xuemeng Song are with the Department of Computer Science and Technology, Shandong University, Qingdao, China (e-mail: bohanhou@foxmail.com; zichaohq@gmail.com; sxmustc@gmail.com).}
\thanks{Haokun Wen is with the School of Computer Science and Technology, Harbin Institute of Technology (Shenzhen), Shenzhen, China. He is also with the Department of Data Science, City University of Hong Kong, Hong Kong, China. (e-mail: whenhaokun@gmail.com).}
\thanks{Meng Liu is with the Department of Computer Science, Shandong Jianzhu University, Jinan, China (e-mail: mengliu.sdu@gmail.com).}
\thanks{Mingzhu Xu is with the School of Software, Shandong University, Jinan, China (e-mail: xumingzhu@sdu.edu.cn).}
\thanks{Corresponding Author: Xuemeng Song and Mingzhu Xu.}

}

\markboth{Hou \MakeLowercase{et al.}: Pseudo Triplet Guided Few-shot Composed Image Retrieval}%
{Hou \MakeLowercase{et al.}: Pseudo Triplet Guided Few-shot Composed Image Retrieval}

\maketitle

\begin{abstract}
Composed Image Retrieval (CIR) is a challenging task that aims to retrieve the target image with a multimodal query, \textit{i.e.}, a reference image, and its complementary modification text.  
As previous supervised or zero-shot learning paradigms all fail to strike a good trade-off between the model's generalization ability and retrieval performance, recent researchers have introduced the task of few-shot CIR (FS-CIR) and proposed a textual inversion-based network based on pretrained CLIP model to realize it. 
Despite its promising performance, the approach encounters two key limitations: simply relying on the few annotated samples for CIR model training and indiscriminately selecting training triplets for CIR model fine-tuning. To address these two limitations, we propose a novel two-stage pseudo triplet guided few-shot CIR scheme, dubbed PTG-FSCIR. 
In the first stage, we propose an attentive masking
and captioning-based pseudo triplet generation method, to construct pseudo triplets from pure image data and use them to fulfill the CIR-task specific pretraining.
In the second stage, we propose a challenging triplet-based CIR fine-tuning method, where we design a pseudo modification text-based sample challenging score estimation strategy and a robust top range-based random sampling strategy for sampling robust challenging triplets to promote the model fine-tuning. 
Notably, our scheme is plug-and-play and compatible with any existing supervised CIR models. 
We test our scheme across two backbones on three public datasets (\textit{i.e.}, FashionIQ, CIRR, and Birds-to-Words), achieving maximum improvements of $\mathbf{13.3\%}$, $\mathbf{22.2\%}$, and $\mathbf{17.4\%}$ respectively, demonstrating our scheme's efficacy.

\end{abstract}
\section{Introduction}
\begin{figure}[t]
  \centering
  \small
  \begin{minipage}{1\linewidth}
    \includegraphics[width=\linewidth]{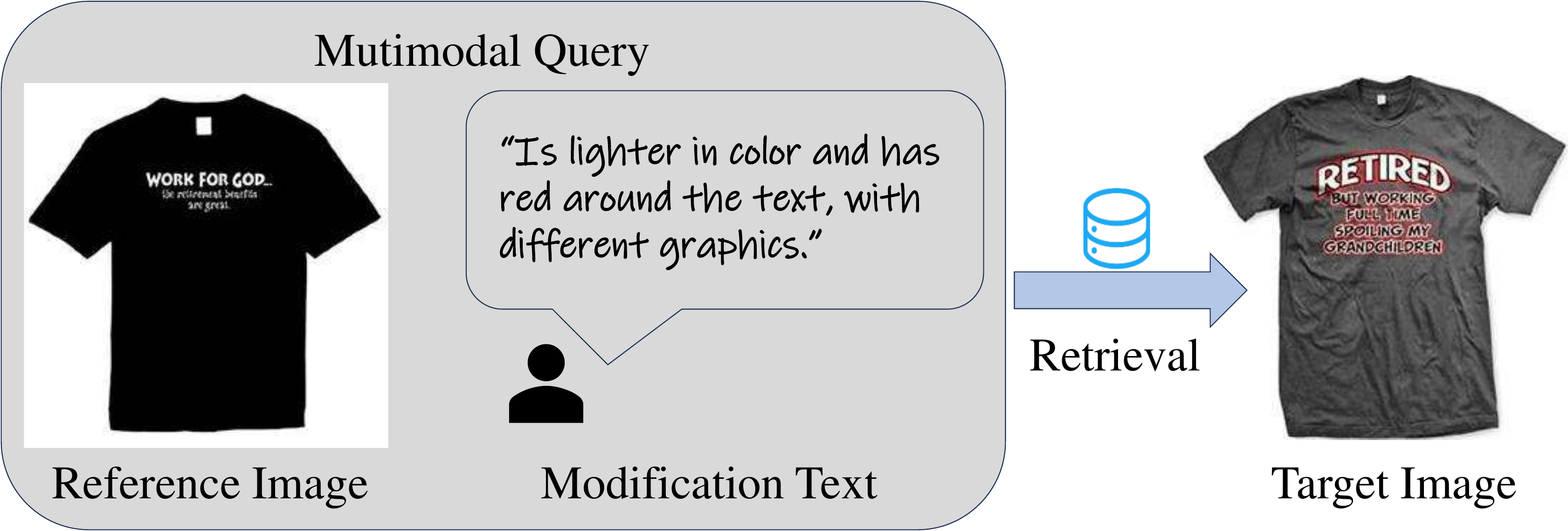}
  \end{minipage}%
  \centering
\caption{ Illustration of the CIR task.}
 \vspace{-2em}
  \label{fig:cir}
\end{figure}

\IEEEPARstart{C}{omposed} Image Retrieval (CIR) is an emerging image retrieval paradigm that allows users to search for the target image flexibly using a multimodal query, which includes a reference image and its corresponding modification text, as shown in Figure~\ref{fig:cir}. 
Due to its promising potential in many real-world areas, such as e-commerce~\cite{r10}, internet retrieval~\cite{r11}, and intelligent robotics~\cite{r12,r13}, CIR has gained significant research attention in recent years. 
Early CIR methods typically fall into the supervised learning paradigm~\cite{r14,r16,r17}. They rely heavily on annotated triplets in the form of  \textless  \emph{reference image, modification text, target image}\textgreater ~during the training phase.
However, annotating the modification text for potential \textless \emph{reference image, target image}\textgreater pairs is time-consuming and laborious, which limits the scale of domain-specific training datasets and constrains the model's generalization ability across various domains. 

To address this limitation,  
several researchers~\cite{r22,r23,r24} have explored Zero-Shot CIR (ZS-CIR), which aims to fulfill CIR without any annotated training triplets. 
Existing ZS-CIR methods can be broadly grouped into two categories: textual inversion-based~\cite{r22,r23,r24} methods and training-free~\cite{r54,r55,r26} methods. The former targets learning a mapping network that can effectively convert the reference image into latent pseudo-word token(s) to unify the multimodal query representation and hence facilitate the target image retrieval with the pretrained vision-language model CLIP~\cite{r28}. Despite their promising results, the learning objectives of these methods focus on the cross-modal transformation, which deviates from the essence of the CIR task, \textit{i.e.}, the multimodal query composition.
Beyond textual inversion-based methods, training-free methods remove the model training process, which typically works on utilizing pretrained image caption generators and large language models (LLMs) to re-summarize the multimodal query with pure text and conduct the target image retrieval with pretrained vision-language models.
Although these methods have achieved compelling success with powerful LLMs, their inference efficiency is inevitably reduced due to the substantial computational resources required by LLMs.

In fact, in real-world scenarios, annotating a small number of samples is often feasible. As a result, a pioneer study~\cite{r27} introduced the few-shot CIR task to bridge the gap between supervised and zero-shot CIR paradigms, and proposed a textual inversion-based FS-CIR model. This model adopts the pretrained CLIP as the feature extraction backbone to fulfill the textual inversion, and incorporates a few randomly selected annotated triplets for training. 
Despite its great success, it falls short in the following two aspects.  
1) \textbf{Simply relying on the few annotated samples for CIR model training}. 
The existing FS-CIR model adopts CLIP as its backbone and simply relies on a few annotated samples to optimize its multimodal query composition capability. This may be inadequate owing to the fact that the modification requirements in the CIR task can be complex and diverse. 
Due to the nature of FS-CIR which can only incorporate a few annotated triplets, we argue that it would be beneficial to conduct the CIR task-specific pretraining with a large amount of unlabeled data to better adapt the model to the CIR task.
2) \textbf{Indiscriminately selecting training triplets for CIR model fine-tuning}. The current model randomly selects triplets, overlooking the fact that different samples contribute differently toward model optimization. An intuitive insight is that more challenging samples—which the model finds difficult to interpret—are instrumental in helping the model master aspects that have not yet been fully learned. Therefore, to maximize the benefits of manual annotation efforts, it is desirable to discriminatory select triplets for annotation. 

To address these limitations, we propose a novel \textbf{P}seudo \textbf{T}riplet \textbf{G}uided \textbf{F}ew-\textbf{S}hot \textbf{CIR} scheme, dubbed PTG-FSCIR\footnote{We will release the code and data to facilitate other researchers.}

PTG-FSCIR involves two stages: pseudo triplet-based CIR pretraining and challenging triplet-based CIR fine-tuning. 
In the first stage, we propose an attentive masking and captioning-based pseudo triplet generation method, to obtain a large number of pseudo triplets from unlabeled pure image data for model pretraining. This stage serves to
equip the model with the foundational knowledge of multimodal query composition for the CIR task.  
The challenging triplet-based CIR fine-tuning stage
targets fine-tuning the pretrained model to enhance its capacity by actively selecting a few challenging samples~\cite{r39,r57,r58}. 
Towards this end,  we first devise a pseudo
modification text-based sample challenging score estimation method, 
where a pseudo modification text is generated for each unlabeled \textless \emph{reference image, target image}\textgreater \ pair, to facilitate its challenging score estimation based on the pretrained model obtained from the first stage. To avoid noisy samples, we further design a top range-based random sampling strategy for obtaining the final challenging samples, where the $3$-$\sigma$ rule~\cite{r53} is used for defining the top range. 

Notably, our PTG-FSCIR scheme can be seamlessly integrated into any existing supervised CIR models (including various networks and loss functions), significantly reducing their reliance on annotated samples while maintaining high performance.

Our main contributions can be summarized as follows:

\begin{itemize}[]
\item We propose a two-stage pseudo triplet guided few-shot CIR scheme. Beyond the existing FS-CIR method, our scheme incorporates the pseudo triplet-based CIR pretraining stage, where an attentive masking and captioning-based pseudo triplet generation method is proposed, for initially optimizing the model's multimodal query composition capability. 
\item  We propose a pseudo
modification text-based sample challenging score estimation method and a top range-based random sampling strategy to improve the quality of the limited annotated samples and hence boost the model's capability in multimodal query composition. As far as we know, we are the first to highlight the quality of the selected training triplets and thereby maximize the benefit brought by these few annotated triplets.  
\item Our scheme is plug-and-play and compatible with various CIR models. We conducted extensive experiments with two cutting-edge CIR models as backbones across three public datasets (\textit{i.e.}, FashionIQ, CIRR, and Birds-to-Words). Across these three datasets, the maximum improvements achieved using our scheme reached $13.3\%$, $22.2\%$, and $17.4\%$, respectively. 
\end{itemize}

\section{Related Work}

\begin{figure*}[t]
  \centering
  
  \begin{minipage}{0.95\linewidth}
    \includegraphics[width=\linewidth]{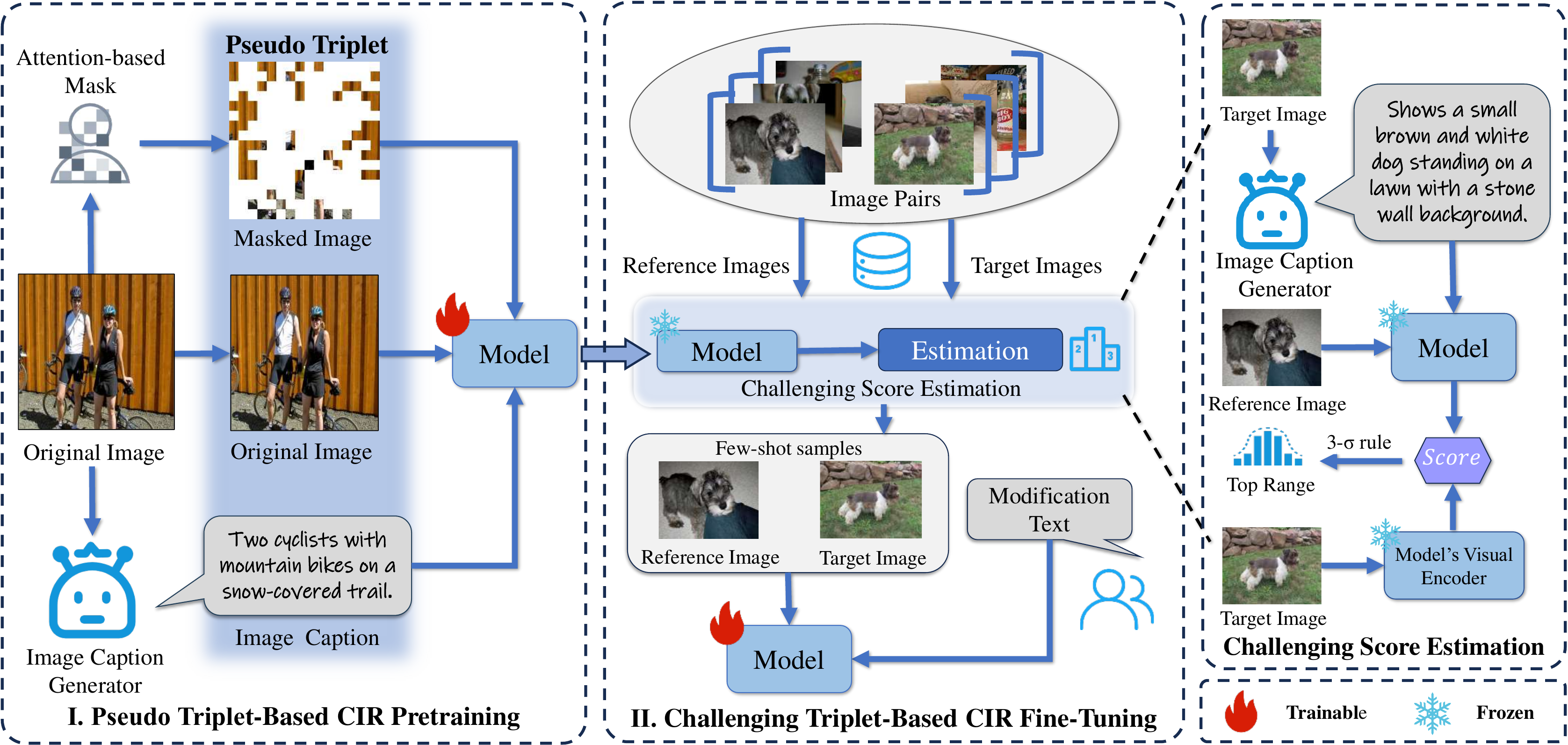}
  \end{minipage}%
  \caption{PTG-FSCIR consists of two stages: pseudo triplet-based CIR pretraining and challenging triplet-based CIR fine-tuning. }
\vspace{-1em}
  \label{fig:arti}
\end{figure*}

\textbf{Composed Image Retrieval.} CIR was initially proposed by Vo et al.~\cite{r2}, allowing users to modify a reference image with text to retrieve desired images. Although existing research efforts have achieved significant success in supervised paradigms~\cite{r14,r16,r17,r19,r21,r2} or zero-shot learning paradigms~\cite{r22,r23,r24}, they all fail to strike a good trade-off between the model’s generalization ability and retrieval performance. Recently, Wu et al.~\cite{r27} introduced the few-shot learning paradigm and proposed a textual inversion-based multimodal query composition strategy to realize FS-CIR. However, they overlooked the fact that relying solely on a limited number of training triplets is insufficient for effectively training the multimodal composition function. In light of this, we introduce the pseudo triplet-based CIR pretraining stage, where we employ an attention-based masked training strategy to generate pseudo triplets from pure image data for pretraining the model, enabling it to acquire primary knowledge related to multimodal query composition.

\textbf{Active Learning.}
Active learning~\cite{r40}, originally introduced to reduce the sample annotation cost for machine learning, aims to actively select the most suitable samples for manually annotating rather than randomly annotating training samples~\cite{r41}, to enhance the model's performance. Currently, active learning 
has been extensively investigated in many research fields, such as image classification~\cite{r57} and semantic segmentation~\cite{r58}. 
However, in the retrieval domain, as far as we know, only Thakur et al.~\cite{r39} have attempted to leverage active learning to tackle the sketch-based image retrieval (SBIR) task. Different from SBIR, CIR involves multimodal queries, making it challenging to select suitable training samples.  
Therefore, in this work, we design the challenging triplet-based CIR fine-tuning stage, where a pseudo modification text-based sample challenging score estimation method and a top range-based random sampling strategy are proposed to effectively realize active learning in CIR. 

\section{Method}
\subsection{Overview}
The goal of FS-CIR is to learn a model that can retrieve target images for a given multimodal query (\textit{i.e.}, a reference image plus a modification text) with only a few annotated triplets for training. Each annotated triplet should be in the form of \textless \emph{reference image, modification text, target image}\textgreater. 
In this work, we propose a novel two-stage pseudo triplet guided few-shot CIR scheme, as shown in Figure~\ref{fig:arti}. It comprises two key stages: pseudo triplet-based CIR pretraining and challenging triplet-based CIR fine-tuning. 
In the first stage, we perform attention-based masking and captioning on images to transform a set of pure images into a set of pseudo triplets in the format of \textless \emph{reference image, modification text, target image}\textgreater \hspace{0.1em} for pretraining our model. We believe that compared to the widely used image-caption pairs, triplet samples are more aligned with the downstream CIR task, and hence benefit the model's knowledge learning regarding multimodal query composition. 
In the second stage, we actively select a small number of challenging \textless \emph{reference image, target image}\textgreater pairs for annotation. Based on the annotated challenging triplets, we fine-tune our pretrained model.

\subsection{Pseudo Triplet-based CIR Pretraining} 
 
The key to the first stage lies in the generation of pseudo triplets. In fact, a previous ZS-CIR work~\cite{r30} has introduced a masking and captioning-based pseudo triplet construction method. This method works on treating a given image as the target image, masking certain patches of the given image to derive the corresponding reference image, and then generating the image caption with the pretrained vision language model BLIP-2)~\cite{r29} to obtain the modification text. The masked image can be regarded as having certain information removed from the original one, while the image caption typically describes the main content of the image. Therefore, the image caption can be regarded as the modification text for restoring the masked information, namely, transforming the masked image back into the original image.
  
However, this work performs random masking, which is likely to yield a reference image with the key subjects incompletely masked. Since the image caption has been regarded as the modification text that describes the main content of the target image, we expect that as many patches related to the main content of the original image as possible should be masked, so that the information in the masked image and the image caption can be complementary to each other.    
 
Towards this end, we propose an attention-based masking strategy, which tends to mask local patches that are more likely to be to the main content of the image.
In particular, to grasp the relationship between each local patch of an image and the main content of the image~\cite{r19,r47}, we adopt a Vision Transformer (ViT) based CLIP visual encoder, where the input is a concatenation of the image global token embedding and local patch embeddings. Specifically, we resort to the attention weights $\mathbf{w} \in \mathbb{R}^{m}$, where $m$ denotes the number of image patches, corresponding to the global features in the last layer of the CLIP visual encoder, which reflects the importance of each local patch within the image. The higher the weight value, the more important the corresponding patch is to the main content of the image.
Formally, we have, 

\begin{equation}
    \left\{
    \begin{aligned}
        &  P_{mask} = \left[{p}_{i}\right], w_i \in {\text{top-}k\left({\mathbf{w}}\right)}, \\
        &\hat{I} = Mask(I,P_{mask}),
    \end{aligned}
    \label{eq:mask_top}
    \right.
\end{equation}
where $P_{mask}$ represents the set of the masked patches, $k$ represents the number of masked patches, ${w}_{i}$ represents the attention weight corresponding to the $i$-th patch ${p}_{i}$, and $\text{top-}k(\mathbf{w})$ denotes the top $k$ entry values of $\mathbf{w}$.$I$ and $\hat{I}$ denote the original image and the masked image, respectively. For image caption generation, we also employ BLIP-2\footnote{Any other advanced image captioning model is applicable.} to derive the caption $\hat{T}$ of the original image. 
Accordingly, we regard \(\langle \hat{I}, \hat{T}, I \rangle\) as a pseudo triplet. In this manner, we can derive a set of pseudo triplets $\mathcal{P}_T$ from the pure image set $\mathcal{S}_I$. Thereafter, we can apply any existing supervised CIR networks with their corresponding loss functions for pretraining.

\subsection{Challenging Triplet-based CIR Fine-tuning}
In the second stage, we aim to fine-tune the model with a few annotated triplet samples. Different from the existing few-shot CIR study~\cite{r27} that randomly selects the annotated samples, we propose to actively select challenging samples for manual annotation and use them for model fine-tuning.   
It is worth noting that existing CIR datasets are primarily constructed by initially gathering the potential \textless \emph{reference image, target image}\textgreater \hspace{0.1em} pairs and then manually annotating the modification text for each image pair. Therefore, the essence of active learning-based CIR fine-tuning lies in how to measure the challenging score of each unlabeled \textless \emph{reference image, target image}\textgreater \hspace{0.1em} pair. 
Towards this end, we propose a pseudo modification text-based sample challenging score estimation method. This method first transforms each candidate \textless \emph{reference image, target image}\textgreater \hspace{0.1em} pair to a pseudo triplet by augmenting the modification text based on the target image caption. It then quantifies the challenging score of this sample based on the feature similarity between the multimodal query and the target image in the pseudo triplet with the help of our pretrained model. 
The underlying philosophy behind adopting the target image caption as the modification text is similar to that in the aforementioned pretraining stage. We believe that the information contained in the target image caption adequately covers the information in the modification text, providing an approximate substitute. 

Specifically, for each \textless \emph{reference image, target image}\textgreater \hspace{0.1em} pair, we first use the image caption generator (BLIP-2)~\cite{r29} for deriving the caption for the target image. Mathematically, let $T_t$ denote the caption of the target image. We then estimate the challenging score $s$ of each query-target image pair according to how well the composed query feature aligns with the target image's feature as follows, 
\begin{equation}
\left\{
\begin{aligned}
f_c &= \phi(I_r, T_t), \\
s &= 1 - \frac{f_c^\top f_t}{\|f_c\| \|f_t\|},
\end{aligned}
\right.
\label{eq:score}
\end{equation}
where \(f_c\) denotes the composed feature vector of the multimodal query and \(f_t\) stands for the feature vector of the target image, both of which are derived by our pretrained model \(\phi\). 
A higher challenging score indicates a greater disparity between the composed query feature and the target image feature, suggesting that the model cannot effectively comprehend the given multimodal query. 
\begin{figure}[t]
  \centering
 \includegraphics[width=0.8\linewidth,height=0.6\linewidth]{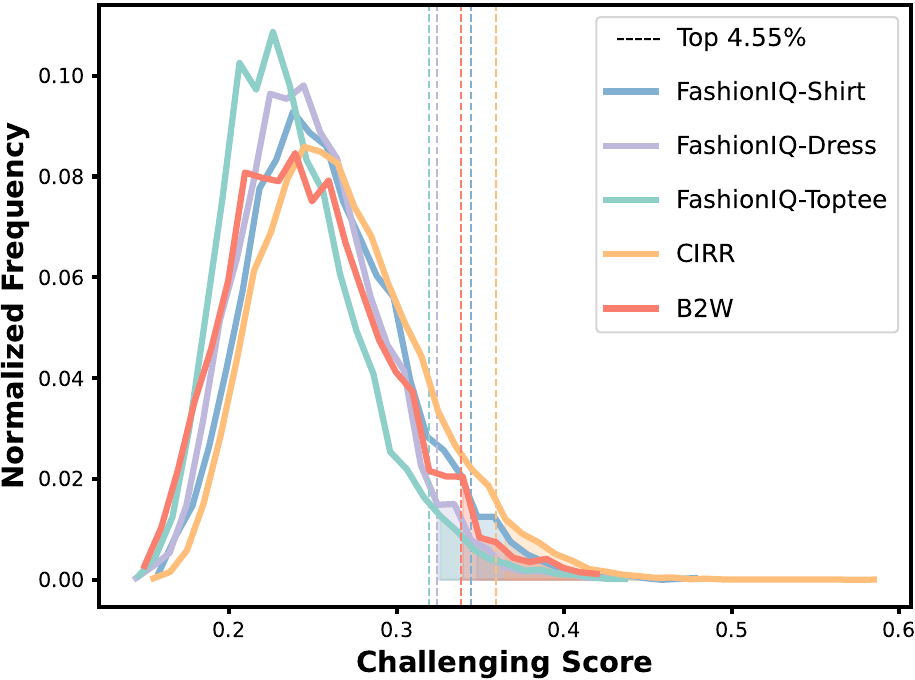}
   \caption{Illustration of skewed challenging score distributions on three subsets of FashionIQ, CIRR, and B2W. The backbone model is SPRC~\cite{r21}.}
  \label{fig:distribution}
  \vspace{-1.3em}
\end{figure}


For challenging sample selection, one naive approach is to directly choose the top-$K$ samples with the highest challenging scores. However, it is possible that the highest-scoring samples are noisy data, which are intrinsic of low quality or yielded by the inaccurate evaluation of our pretrained model. To address these noisy cases, instead of directly selecting the top-$K$ samples, we propose a top range-based random sampling strategy. This strategy first treats the challenging scores of all candidate triplets as a distribution and defines a candidate range following the $3$-$\sigma$ rule, commonly used to identify samples that are statistically significant~\cite{r53}. Ultimately, the proposed strategy randomly selects $K$ triplets from the candidate range as the final selected challenging triplets. 

Specifically, considering that the range beyond $3$-$\sigma$ is too narrow, and the range beyond $1$-$\sigma$ is too broad, we opt for the $2$-$\sigma$ standard. According to the original $2$-$\sigma$ standard, which is defined for a symmetric distribution, we should take the range that falls into the top and bottom $2.75\%$ of the distribution, corresponding to the areas beyond ±$2\sigma$, as the candidate range. However, in our context, since the pretrained model \(\phi\) possesses foundational knowledge of multimodal query composition from the first pretraining stage, it tends to correctly handle a substantial portion of the above pseudo triplets, \textit{i.e.}, assigning lower scores for all triplets. In other words, the challenging score distribution for all candidate pseudo triplets tends to be skewed towards lower values, as shown in Figure~\ref{fig:distribution}. Therefore, instead of selecting $2.75\%$ from each tail following the conventional $2$-$\sigma$ standard, we select the top $4.55\%$ of the most challenging sample to ensure that a sufficient number of significant samples can be captured within a skewed distribution.

\section{Experiment}
\begin{table*}[t]
    \centering
    \caption{Performance comparison on FashionIQ dataset with respect to R@$k$(\%). The best results are highlighted in boldface. The backbone model equipped with our PTG scheme is highlighted by employing a gray background.
}
    
    \vspace{-1em}
    \resizebox{1.0\linewidth}{!}{
   \begin{tabular}{lcccccccc}
    
    \hline
        \multirow{2}{*}{Method} & \multicolumn{4}{c}{R@$10$ ~ ~ ~}& \multicolumn{4}{c}{R@$50$ ~ ~ ~}\\ 
        \cmidrule(lr){2-5}
    \cmidrule(lr){6-9}
         ~ & $2$ & $4$ & $8$ & $16$ & $2$ & $4$ & $8$ & $16$ \\ \hline
    DCNet~\textcolor{gray}{(AAAI'21)}& $1.9{\pm}0.0$ & $2.1{\pm}0.2$ & $2.3{\pm}0.1$ & $2.9{\pm}0.1$ & $9.3{\pm}3.6$ & $10.4{\pm}4.1$ & $10.9{\pm}4.0$ & $12.6{\pm}3.9$\\
       ComposeAE~\textcolor{gray}{(WACV'21)}& $2.8{\pm}0.0$ & $3.1{\pm}0.2$ & $3.5{\pm}0.2$ & $4.0{\pm}0.1$ & $9.2{\pm}3.6$ & $10.4{\pm}4.1$ & $10.9{\pm}4.0$ & $12.2{\pm}3.9$\\
       TG-CIR~\textcolor{gray}{(MM'23)} & $6.8{\pm}0.3$ & $7.1{\pm}0.4$ & $7.3{\pm}0.4$ & $8.7{\pm}0.3$ & $15.4{\pm}0.7$ & $16.0{\pm}0.6$ & $18.2{\pm}0.6$ & $19.2{\pm}0.7$ \\
       DQU-CIR~\textcolor{gray}{(SIGIR'24)} &$18.1{\pm}1.3$ & $19.3{\pm}1.1$ & $19.7{\pm}0.7$ & $20.8{\pm}0.8$ & $34.2{\pm}1.1$ & $35.0{\pm}0.9$ & $35.3{\pm}0.6$ & $38.4{\pm}0.9$ \\ \hline
       PromptCLIP~\textcolor{gray}{(AAAI'23)}& $18.9{\pm}0.6$ & $19.8{\pm}0.8$ & $19.9{\pm}0.7$ & $21.2{\pm}0.6$ & $36.9{\pm}1.2$ & $38.5{\pm}1.1$ & $38.4{\pm}0.6$ & $40.4{\pm}0.7$
\\ \hline

CLIP4CIR~\textcolor{gray}{(CVPR'22)} & $19.0{\pm}0.5$ & $22.2{\pm}0.4$& $23.3{\pm}1.3$& $25.0{\pm}0.4$& $37.9{\pm}0.5$ & $39.4{\pm}0.5$& $40.3{\pm}1.0$& $44.0{\pm}0.8$\\ 
\rowcolor{gray!20}
  $\text{CLIP4CIR}_{+PTG}$ & $28.1{\pm}0.5$& $29.1{\pm}0.4$& $30.5{\pm}0.4$& $\mathbf{31.6}{\pm}0.5$& $47.8{\pm}0.4$& $48.5{\pm}0.5$& $51.1{\pm}0.6$& $51.8{\pm}0.8$\\ 
  Improvement & $\uparrow9.1\%$ & $\uparrow6.9\%$ & $\uparrow7.2\%$ & $\uparrow6.6\%$ & $\uparrow9.9\%$ & $\uparrow9.1\%$ & $\uparrow10.8\%$ & $\uparrow7.8\%$\\ \hline

SPRC~\textcolor{gray}{(ICLR'24)} & $19.6{\pm}0.5$& $23.5{\pm}0.1$& $25.7{\pm}0.5$& $26.3{\pm}0.7$& $37.9{\pm}0.1$& $42.1{\pm}0.1$& $46.5{\pm}0.5$& $47.6{\pm}0.8$\\
\rowcolor{gray!20}
  $\text{SPRC}_{+PTG}$ & $\mathbf{29.1}{\pm}0.3$& $\mathbf{29.8}{\pm}0.4$& $\mathbf{30.6}{\pm}0.5$& $31.3{\pm}0.6$& $\mathbf{51.2}{\pm}0.4$& $\mathbf{51.7}{\pm}0.4$& $\mathbf{51.9}{\pm}0.7$ &$\mathbf{52.4}{\pm}0.6$\\  
  Improvement & $\uparrow9.5\%$ & $\uparrow6.3\%$ & $\uparrow4.9\%$ & $\uparrow5.0\%$ & $\uparrow13.3\%$ & $\uparrow9.6\%$ & $\uparrow5.4\%$ & $\uparrow4.8\%$
\\ \hline
\end{tabular} 
}

\label{table1}
\end{table*}

\begin{table*}[t]
     \begin{center}
    \caption{Performance comparison on CIRR dataset with respect to R@$k$(\%). The best results are highlighted in boldface. 
The backbone model equipped with our PTG scheme is highlighted by employing a gray background.
}
\vspace{-1em}
\resizebox{1.0\linewidth}{!}{
    \begin{tabular}{lcccccccc}
   
    \hline
        \multirow{2}{*}{Method} & \multicolumn{4}{c}{R@$1$ ~ ~ ~}& \multicolumn{4}{c}{R@$5$ ~ ~ ~}\\ 
        \cmidrule(lr){2-5}
    \cmidrule(lr){6-9}
         ~ & $2$ & $4$ & $8$ & $16$ & $2$ & $4$ & $8$ & $16$ \\ \hline
     DCNet~\textcolor{gray}{(AAAI'21)}& $1.5{\pm}0.0$ & $2.1{\pm}0.2$ & $2.4{\pm}0.3$ & $3.2{\pm}0.2$ & $5.2{\pm}0.3$ & $7.7{\pm}1.4$ & $8.9{\pm}0.8$ & $10.1{\pm}0.4$\\
       ComposeAE~\textcolor{gray}{(WACV'21)}& $1.3{\pm}0.1$ & $2.6{\pm}0.0$ & $2.9{\pm}0.1$ & $2.6{\pm}0.2$ & $3.3{\pm}0.2$ & $5.9{\pm}0.1$ & $6.3{\pm}0.4$ & $8.6{\pm}0.6$\\
       TG-CIR~\textcolor{gray}{(MM'24)} & $7.3{\pm}0.3$ & $7.9{\pm}0.6$ & $8.6{\pm}0.4$ & $8.5{\pm}0.7$ & $26.0{\pm}0.4$ & $27.6{\pm}0.9$ & $27.4{\pm}0.5$ & $27.9{\pm}0.8$\\
        DQU-CIR~\textcolor{gray}{(SIGIR'24)} &$13.8{\pm}0.8$ & $14.1{\pm}1.1$ & $14.4{\pm}0.5$ & $14.5{\pm}0.8$ & $38.2{\pm}0.9$ & $39.5{\pm}0.8$ & $40.8{\pm}1.1$ & $40.4{\pm}1.3$ \\ \hline
       PromptCLIP~\textcolor{gray}{(AAAI'23)}& $13.2{\pm}0.5$ & $14.1{\pm}0.5$ & $15.8{\pm}0.3$ & $16.0{\pm}0.4$ & $34.6{\pm}0.3$ & $36.7{\pm}0.4$ & $41.0{\pm}0.6$ & $40.8{\pm}0.9$\\ \hline

CLIP4CIR~\textcolor{gray}{(CVPR'22)} & $15.9{\pm}0.2$ & $18.6{\pm}0.3$ & $23.1{\pm}0.3$ & $24.4{\pm}0.3$ & $41.7{\pm}0.4$ & $47.3{\pm}0.3$ & $52.0{\pm}0.3$ & $53.5{\pm}0.6$\\ 
\rowcolor{gray!20}
$\text{CLIP4CIR}_{+PTG}$& $28.2{\pm}0.3$ & $28.7{\pm}0.3$ & $30.4{\pm}0.6$ & $33.4{\pm}0.3$ & $59.9{\pm}0.4$ & $59.6{\pm}0.6$ & $62.5{\pm}0.4$ & $64.2{\pm}0.5$\\ 
Improvement & $\uparrow12.3\%$ & $\uparrow10.1\%$ & $\uparrow7.3\%$ & $\uparrow9.0\%$ & $\uparrow18.2\%$ & $\uparrow12.3\%$ & $\uparrow10.5\%$ & $\uparrow10.7\%$ \\  \hline
SPRC~\textcolor{gray}{(ICLR'24)} & $18.2{\pm}0.4$ & $25.2{\pm}0.3$ & $27.1{\pm}0.5$ & $32.6{\pm}1.1$ & $40.1{\pm}0.6$ & $52.3{\pm}0.7$ & $55.2{\pm}0.3$ & $61.3{\pm}0.8$\\ 
\rowcolor{gray!20}
$\text{SPRC}_{+PTG}$ & $\mathbf{33.9}{\pm}0.5$ & $\mathbf{34.9}
{\pm}0.6$ & $\mathbf{35.7}{\pm}0.6$ & $\mathbf{38.6}{\pm}0.7$ & $\mathbf{62.3}{\pm}0.4$ & $\mathbf{63.5}{\pm}0.4$ & $\mathbf{65.5}{\pm}0.2$ & $\mathbf{67.5}{\pm}0.4$\\  
Improvement & $\uparrow15.7\%$ & $\uparrow9.7\%$ & $\uparrow8.6\%$ & $\uparrow6.0\%$ & $\uparrow22.2\%$ & $\uparrow11.2\%$ & $\uparrow10.3\%$ & $\uparrow6.2\%$ \\ \hline
\end{tabular} }

\label{table2}  
\end{center}
\end{table*}

\subsection{Experiment Settings}
\textbf{Datasets.}
To evaluate the performance of our scheme in various CIR tasks, we employed three public datasets for evaluation: FashionIQ~\cite{r4}, CIRR~\cite{r1}, and Birds-to-Words~\cite{r45}. 
\textbf{FashionIQ} contains fashion items of three categories: Dresses, Shirts, and Tops\&Tees. Approximately 46K images and 15K images are used for training and testing, respectively. Finally, there are $18$K triplets for training and $6$K for testing.
\textbf{CIRR} comprises 21K real-life open-domain images taken from the $\text{NLVR}^2$ dataset~\cite{r49}. It contains over $36,000$ pairs of real-world triplets data, where $80\%$ for training, $10\%$ for validation, and $10\%$ for testing. 
\textbf{Birds-to-Words} (B2W) consists of $3.5$k images of birds, where $2.7$K training image pairs and $0.6$K testing image pairs are constructed. Each image pair is tagged with a descriptive comparison between them. 
Based on the construction process of image pairs, triplets in B2W can be divided into the following seven categories: base, visual, sameSpecies, sameGenus, sameFamily, sameOrder, and sameClass. 

\textbf{Implementation Details.}
As our scheme can be applied to arbitrary supervised CIR models, we selected two advanced and representative supervised CIR models, including CLIP4CIR~\cite{r16}
and SPRC~\cite{r21} as backbone models. The main ideas of these two models are different, where CLIP4CIR is a CLIP feature fusion-based model and SPRC is a text inversion-based model. This diverse selection underpins a robust evaluation of our scheme's effectiveness and generalization. 

In the first stage, we utilized the ImageNet~\cite{r46} test set as our pretraining dataset due to its rich and diverse collection of images~\cite{r23}.
We employed the BLIP-2-opt-$2.7$B model to generate image captions as modification texts. 
Considering that a higher masking rate can significantly enhance performance~\cite{r30,r31}, we segmented each image into $256$ patches, \textit{i.e.} $m=256$, and set the number of masked patches $k$ in Eqn. ($\ref{eq:mask_top}$) to 192,  to keep the masking ratio being $75\%$.
Additionally, we set our learning rate and optimizer according to the specifications mentioned in the backbone methods to ensure optimal training conditions. 
In the second stage, regarding the number of triplet samples to be chosen,  following the previous FS-CIR study~\cite{r27}, we conducted few-shot training with $2$, $4$, $8$, and $16$ shots per category for all datasets. 
Notably, unlike FashionIQ and B2W, the CIRR dataset originates from a diverse open-world collection and lacks explicit categorization. Therefore, we employed the $k$-means clustering algorithm to generally group CIRR into {four categories}. This setting is to make the size of each category in CIRR similar to that in FashionIQ, the total volume of which shares a similar scale with that of CIRR. Learning rates are set to $1e-5$ for FashionIQ and B2W, and $5e-5$ for CIRR. Following~\cite{r27}, we repeated each experiment five times and reported the mean and standard error. 

\textbf{Evaluation Metrics.}
Following previous work~\cite{r23,r27,r21}, we adhered to the standard evaluation protocols for each dataset and utilized the Recall@$k$ (R@$k$) as the evaluation metric. 
Similar to the existing FS-CIR study~\cite{r27}, for the FashionIQ dataset and B2W dataset,  we reported the average results across all their subsets for R@$10$ and R@$50$.Notably, for the CIRR dataset, we reported results for R@$1$ and R@$5$.

\begin{table*}[t]
    \centering
    
\caption{Performance comparison on B2W dataset with respect to R@$k$(\%). The best results are highlighted in boldface. The backbone model equipped with our PTG scheme is highlighted by employing a gray background.
}
    \resizebox{1.0\linewidth}{!}{
    \begin{tabular}{lcccccccc}
    
    \hline
        \multirow{2}{*}{Method} & \multicolumn{4}{c}{R@$10$ ~ ~ ~}& \multicolumn{4}{c}{R@$50$ ~ ~ ~}\\ 
        \cmidrule(lr){2-5}
    \cmidrule(lr){6-9}
         ~ & $2$ & $4$ & $8$ & $16$ & $2$ & $4$ & $8$ & $16$ \\ \hline
       DCNet~\textcolor{gray}{(AAAI'21)}& $2.1{\pm}0.2$ & $1.9{\pm}0.3$ & $2.4{\pm}0.2$ & $3.0{\pm}0.2$ & $9.4{\pm}1.6$ & $10.4{\pm}1.2$ & $12.6{\pm}0.4$ & $13.9{\pm}0.5$\\
       ComposeAE~\textcolor{gray}{(WACV'21)}& $1.7{\pm}0.1$ & $2.2{\pm}0.2$ & $2.5{\pm}0.2$ & $3.1{\pm}0.4$ & $9.6{\pm}0.7$ & $10.0{\pm}0.7$ & $12.9{\pm}1.1$ & $13.6{\pm}0.9$\\
       TG-CIR~\textcolor{gray}{(MM'23)}& $14.3{\pm}0.4$ & $18.1{\pm}0.3$ & $24.1{\pm}0.4$ & $29.3{\pm}0.5$ & $37.4{\pm}0.4$ & $43.6{\pm}0.3$ & $55.4{\pm}0.4$ & $58.2{\pm}0.7$
 \\  \hline
       PromptCLIP~\textcolor{gray}{(AAAI'23)}& $9.5{\pm}0.4$ & $11.0{\pm}0.6$ & $12.9{\pm}0.7$ & $14.2{\pm}0.6$ & $29.3{\pm}0.8$ & $36.1{\pm}0.5$ & $38.7{\pm}0.8$ & $39.2{\pm}0.4$ \\ \hline

CLIP4CIR~\textcolor{gray}{(CVPR'22)} & $34.1{\pm}0.7$ & $39.4{\pm}0.6$ & $41.3{\pm}0.3$ & $42.8{\pm}0.3$ & $55.9{\pm}0.4$ & $62.3{\pm}0.5$ & $62.9{\pm}0.6$ & $63.0{\pm}0.5$\\ 
\rowcolor{gray!20}
$\text{CLIP4CIR}_{+PTG}$ & $\mathbf{44.8}{\pm}0.5$ & $\mathbf{45.6}{\pm}0.4$ & $\mathbf{46.6}{\pm}0.4$ & $\mathbf{48.5}{\pm}0.3$ & $\mathbf{66.1}{\pm}0.5$ & $\mathbf{65.3}{\pm}0.6$ & $\mathbf{69.7}{\pm}0.3$ & $\mathbf{69.9}{\pm}0.4$ \\
Improvement & $\uparrow10.7\%$ & $\uparrow6.2\%$ & $\uparrow5.3\%$ & $\uparrow5.7\%$ & $\uparrow10.2\%$ & $\uparrow3.0\%$ & $\uparrow6.8\%$ & $\uparrow6.9\%$\\

\hline
SPRC~\textcolor{gray}{(ICLR'24)} & $15.8{\pm}0.5$ & $23.7{\pm}0.5$ & $25.9{\pm}0.6$ & $29.8{\pm}0.7$ & $44.3{\pm}0.8$ & $53.1{\pm}0.6$ & $54.1{\pm}0.7$ & $55.0{\pm}0.9$ 
\\ 
\rowcolor{gray!20}
$\text{SPRC}_{+PTG}$ & $33.2{\pm}1.4$ & $35.2{\pm}0.8$ & $40.7{\pm}0.8$ & $43.4{\pm}1.5$ & $58.5{\pm}0.8$ & $62.1{\pm}0.9$ & $64.3{\pm}0.8$ & $66.1{\pm}1.1$ \\
Improvement & $\uparrow17.4\%$ & $\uparrow11.5\%$ & $\uparrow14.8\%$ & $\uparrow13.6\%$ & $\uparrow14.2\%$ & $\uparrow9.0\%$ & $\uparrow10.2\%$ & $\uparrow11.1\%$
\\  \hline
\end{tabular} }

\label{table3}

\end{table*}

\begin{table*}[t]
    \centering
       \caption{Ablation study results on three datasets with respect to 
R@$k$(\%). 
``Stage1'' and ``Stage2'' refer to the setting of the pretraining stage and fine-tuning stage, respectively.
The best results are highlighted in boldface.}
    
     \resizebox{1.0\linewidth}{!}{
    \begin{tabular}{c|c|c|cc|cc|cc}
    
    \hline
     \multirow{2}{*}{AM\#} & \multirow{2}{*}{Stage1} & \multirow{2}{*}{Stage2} & \multicolumn{2}{c|}{FashionIQ} & \multicolumn{2}{c| }{B2W} & \multicolumn{2}{c}{CIRR} \\
    \cline{4-9}
    & & & R@$10$ & R@$50$ & R@$10$ & R@$50$ & R@$1$ & R@$5$\\
    \hline \hline

    \cline{1-9}
                           
                         1 &No-Pretrain & Random & $26.3{\pm}0.7$ & $47.6{\pm}0.8$ & $29.8{\pm}0.7$ & $55.0{\pm}0.9$ & $32.6{\pm}1.1$ & $61.3{\pm}0.8$ \\
                         
                          2 &Random-Mask& 2-$\sigma$  & $31.1{\pm}0.7$ & $51.9{\pm}1.1$ & ${41.4}{\pm}1.2$ & ${65.3}{\pm}1.1$ & ${36.4}{\pm}0.6$ &
                          ${66.1}{\pm}0.4$\\
                        \cdashline{1-9}[1pt/1pt]
                       
                         3 &\multirow{3}{*}{Attention-Mask}
                        & Random & $30.5{\pm}0.6$ & $51.7{\pm}0.9$ &
                          $38.8{\pm}0.8$ & $63.4{\pm}0.7$ &
                          $36.5{\pm}0.5$ &
                          $65.1{\pm}0.6$\\ 
                             
                           4 & &Top-$K$ &$28.3{\pm}0.5$  & $48.6{\pm}0.4$  & $39.9{\pm}0.9$ &  $63.1{\pm}1.1$  &$35.8{\pm}0.6$  & $64.1{\pm}0.5$ \\
                           5 & & Easy & $29.9{\pm}0.2$ & $51.5{\pm}0.8$ & $38.5{\pm}0.4$ & $61.2{\pm}1.4$ & 
                          $36.2{\pm}0.8$&
                          $64.7{\pm}0.8$\\ 
                           \hline
                           \rowcolor{gray!20}
                          \multicolumn{3}{c|}{$\textbf{SPRC}_{+PTG}$ } & $\mathbf{31.3}{\pm}0.6$ & $\mathbf{52.4}{\pm}0.8$ & $\mathbf{43.4}{\pm}1.5$ & $\mathbf{66.1}{\pm}1.1$ & $\mathbf{38.6}{\pm}0.7$ &
                          $\mathbf{67.5}{\pm}0.4$\\
                          
    \hline
    \end{tabular}}
 
    \label{table4}
\end{table*}
\subsection{Performance Comparison}
As aforementioned, FS-CIR is a new challenging task, and there is only one FS-CIR model, \textit{i.e.},  PromptCLIP~\cite{r27} can be used for comparison. Therefore, 
to validate the effectiveness of our scheme,  following~\cite{r27},  we also adopted four representative supervised CIR models, including DCNet~\cite{r50}, ComposeAE~\cite{r51}, TG-CIR~\cite{r19}, and DQU-CIR~\cite{r56} \footnote{DQU-CIR does not provide ``keywords'' data for training on B2W,  here we only present the results on two other datasets.} for comparison. Notably, in the context of FS-CIR, these supervised CIR models are trained with only a few annotated samples.  

Tables~\ref{table1}-~\ref{table3} summarize the performance comparison on the three datasets. From these tables, we obtained the following observations. 
1) In a series of baselines, methods with pretrained models (\textit{i.e.}, TG-CIR, DQU-CIR, and PromptCLIP) perform better. It is noteworthy that the composition function of DQU-CIR is very similar to that of CLIP4CIR, but DQU-CIR, by integrating the multimodal query at the raw-data level, achieves performance far superior to CLIP4CIR in supervised mode, yet does not perform as well in a few-shot setting. These observations all reflect the role of pretraining in handling complex multimodal query composition. 
2) Our scheme with both backbones (\textit{i.e.}, CLIP4CIR and SPRC) consistently outperforms the FS-CIR method PromptCLIP, which reflects the effectiveness of our scheme for FS-CIR.
3) Although the two backbones are advanced supervised CIR methods, their performance is unsatisfactory in the few-shot CIR context, where only a few randomly annotated samples are available. Nevertheless, when equipped with our scheme, both backbones show significant performance improvement. This indicates the necessity of conducting the pseudo triplet-based pretraining and challenging triplet-based fine-tuning.

\subsection{Ablation Study} 
To verify the influence of each key component in our scheme, we conducted ablation study experiments on three datasets, where  SPRC is used as the model backbone.

\begin{itemize}
    \item \textbf{Pseudo triplet-based CIR pretraining}. 
    To gain deep insights into this stage, we introduced two variants: removing the pretraining stage (\textbf{AM1}) and adopting the random masking strategy for pseudo triplet-based pretraining (\textbf{AM2}). Notably, considering that without pretraining, the challenging samples obtained in the second stage can be rather noisy, we randomly selected samples for fine-tuning the model in \textbf{AM1}.

    \item \textbf{Challenging triplet-based CIR fine-tuning}. 
    For this stage, to justify the overall active learning strategy, we randomly selected samples for the model fine-tuning (\textbf{AM3}); to study the effectiveness of our top range-based sampling strategy, we introduced two variants: directly selecting the top-$K$ samples with the highest challenging scores for model fine-tuning  (\textbf{AM4}) and sampling easy triplets from the bottom $25$\%\footnote{We empirically found our pretrained model from the first stage can accurately retrieve around 25\% of the data on all three datasets.} of scores calculated by Eqn.~\ref{eq:score}  (\textbf{AM5}).
    
\end{itemize}

Owing to the limited space, here we only reported the results under the 16-shot scenario in Table~\ref{table4}. 
From Table~\ref{table4}, we gained the following observations.
1) \textbf{AM3} significantly outperforms \textbf{AM1}, which proves that the pseudo triplet-based CIR pretraining stage does promote the model's multimodal query composition capability.
2) \textbf{AM2} performs worse than $\text{SPRC}_{+PTG}$. This is reasonable as, compared to random masking, our attention-based masking strategy exhibits a more profound exploration of the complementary relationships between image-text pairs. Therefore, the generated pseudo triplets more accurately emulate real-world triplets, thereby providing the CIR model with better primary knowledge for multimodal query composition.
3) $\mathbf{\textbf{SPRC}_{+PTG}}$ demonstrates superior performance over \textbf{AM3-5}. This not only underscores the importance of active learning-based sample selection in the few-shot learning paradigm but also validates the effectiveness of our 2-$\sigma$ standard-guided top range-based sampling strategy. 
4) \textbf{AM4} performs worse than \textbf{AM5}. The possible explanation is that samples with the highest challenging scores may be too hard to interpret and noisy, which could not benefit the model fine-tuning. 5) \textbf{AM5} performs worse than \textbf{AM3}. This may be because easy samples are more likely to have already been mastered in the pretraining stage. In contrast, \textbf{AM3} includes more challenging samples compared to \textbf{AM5}, which are beneficial for further optimization of the model.

\subsection{Performance Comparison in ZS-CIR} 
Although our model is proposed for FS-CIR, the model with only the pretraining stage can be applied to solve the task of ZS-CIR.  Therefore, for comprehensive comparison, we also compare our models trained simply by our pretraining stage with seven state-of-the-art ZS-CIR methods, including five textual inversion based ZS-CIR methods
(\textit{i.e.}, Pic2Word~\cite{r22}, SEARLE~\cite{r23}, ContextI2W~\cite{r24}, FIT4CIR~\cite{r47}, LinCIR~\cite{r59}) and two other generated triplets based ZS-CIR methods (\textit{i.e.} CompoDiff~\cite{r18}, and MCL~\cite{r60} with its various LLM configurations.) All methods use the ViT-L configuration for their visual encoders. It is worth mentioning that CompoDiff was trained with approximately $18$M  triplet data points generated by text-prompt diffusion models, while MCL used approximately $2.7$M pseudo triplets in the form of \textless \emph{reference image, generated modification text, generated target caption}\textgreater \ produced by LLM. In comparison, our pretraining data consists of only $100$K triplets generated by attention-based masking.

As can be seen from Table~\ref{tab:cirr_test_exp}, 
although our scheme is not specifically designed for ZS-CIR, it still holds its ground against current state-of-the-art ZS-CIR methods. This reconfirms the benefits of conducting the CIR-task specific pretraining with pseudo triplets generated by our attention-based masking strategy in endowing the model with fundamental knowledge of multimodal query composition for the CIR task. 
Notably, compared to other generated triplets based ZS-CIR methods that require larger datasets for training, our method still performs exceptionally well. This highlights the superiority of our attention-based masked training strategy in ZS-CIR, which significantly reduces the cost of data construction and improves training efficiency.

\begin{table*}[htbp]
  \centering \small
  \caption{Performance comparison with the ZS-CIR methods on CIRR dataset. The best results are highlighted in boldface. The backbone models equipped with our first stage pretraining scheme are highlighted by employing a gray background. – denotes results not reported in the original paper.}
   \resizebox{0.93\linewidth}{!}{ 
   \begin{tabular}{llccccccc} 
    \toprule
    \multicolumn{1}{l}{} & \multicolumn{1}{c}{} & \multicolumn{4}{c}{Recall} & \multicolumn{3}{c}{Recall$_{\text{Subset}}$} \\
    \cmidrule(lr){3-6}
    \cmidrule(lr){7-9}

    \multicolumn{1}{l}{Type} & \multicolumn{1}{l}{Methods} & R@$1$ & R@$5$ & R@$10$ & R@$50$ & R@$1$ & R@$2$ & R@$3$\\ 
   \midrule
 \multirow{5}{*}{Textual inversion}
 &Pic2Word{~\textcolor{gray}{(CVPR'23)}} & $23.9$ & $51.7$ & $65.3$ & $87.8$  & - & - & - \\
&SEARLE-XL-OTI{~\textcolor{gray}{(ICCV'23)}} & $24.9$ & $52.3$ & $66.3$ & $88.6$ & $53.8$ &$74.3$ &$86.9$ \\
    &ContextI2W{~\textcolor{gray}{(AAAI'24)}} & $25.6$ & $55.1$ & $65.3$ & $87.8$  & - & - & - \\
    &FTI4CIR{~\textcolor{gray}{(SIGIR'24)}} & $25.9$ & $55.6$ & $67.7$ & $89.7$  & $55.2$ & $75.9$ & $88.0$ \\
    & LinCIR{~\textcolor{gray}{(CVPR'24)}} & $25.0$ & $53.3$ & $66.7$ & $-$ & $57.1$ & $77.4$ & $88.9$ \\

 \midrule

 \multirow{6}{*}{Generated triplets}
 & CompoDiff{~\textcolor{gray}{(TMLR'24)}} & $19.4$ & $53.8$ &$72.0$ & $90.9$ & $\mathbf{64.5}$ & $78.8$  & $89.3$ \\
 & MCL-(OPT-2.7B){~\textcolor{gray}{(ICML'24)}} & $23.3$ & $54.2$ & $67.2$ & $90.1$ & $58.2$ & $79.4$ & $90.5$ \\
 & MCL-(OPT-6.7B){~\textcolor{gray}{(ICML'24)}} & $24.2$ & $56.0$ & $69.2$ & $90.8$ & $59.5$ & $80.3$ & $91.1$ \\
 & MCL-(LLaMA2-7B){~\textcolor{gray}{(ICML'24)}} & $26.2$ & $56.8$ & $70.0$ & $91.4$ & $61.5$ & $\mathbf{81.6}$ & $91.9$ \\
\rowcolor{gray!20} \cellcolor{white}& CLIP4CIR(\text{Our\_Pretrain}) & $25.9$ & $\mathbf{57.3}$ & $\mathbf{73.6}$ & $ 90.9$ & $52.6$ & $74.4$ & $87.1$\\  
\rowcolor{gray!20} \cellcolor{white}& SPRC(\text{Our\_Pretrain}) & $\mathbf{26.4}$ & $56.9$ &$69.9$ & $\mathbf{91.6}$ & $55.9$ & $79.1$  & $\mathbf{92.0}$ \\

\bottomrule
    \end{tabular} }

   \label{tab:cirr_test_exp}
 \end{table*}

\subsection{Sensitivity Analysis on Masking Rate}
\begin{figure}[htbp]
  \centering
  \begin{minipage}{0.8\linewidth}
    \includegraphics[width=\linewidth]{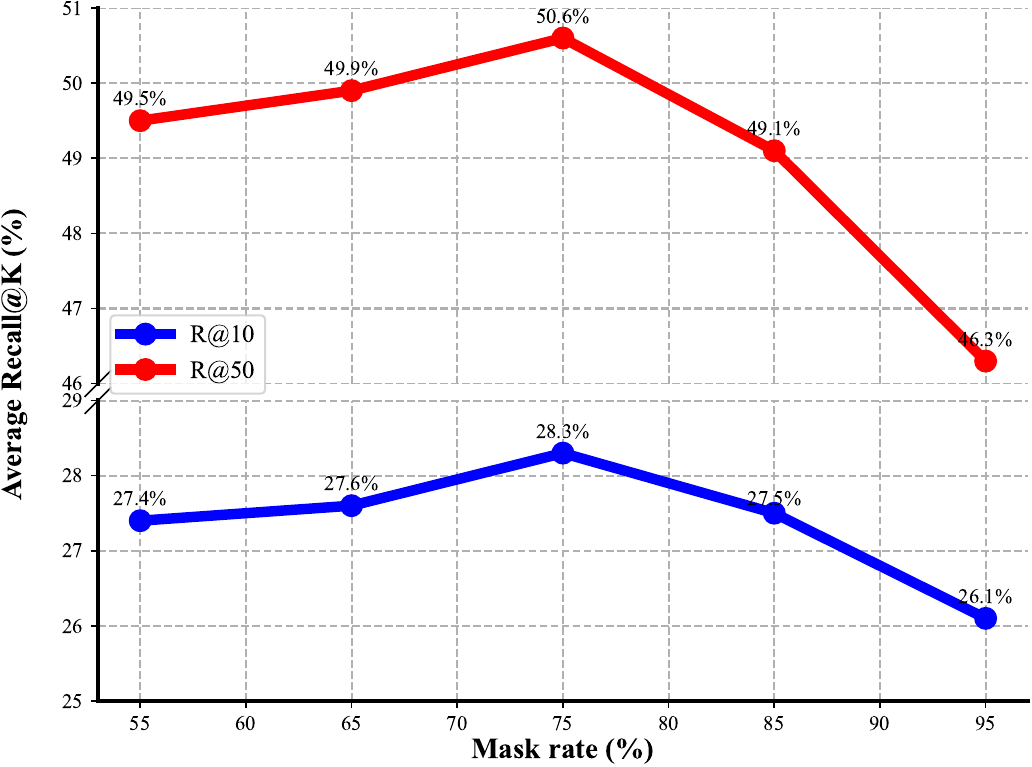}
  \end{minipage}%
\caption{Sensitivity experiments on the masking rate, with SPRC as the backbone on FashionIQ, the red line represents Average R@10, and the blue line represents Average R@50.}
\vspace{-1.5em}
  \label{fig:Sensitivity}
\end{figure}
In this part, we tested the sensitivity of our scheme regarding the masking rate $rate_m$ for the stage of attention-based mask pretraining ($rate_m =  k/m$).
In particular, we varied the masking rates $rate_m$ from $55$\% to $95\%$ at the step of $10\%$ and with the number of patches rounded up. Notably, to clearly demonstrate the performance under different masking rates, we report the results of our model with only the first pretraining stage. 
Figure~\ref{fig:Sensitivity} shows the performance of our model with different masking rates on the FashionIQ dataset, where SPRC is adopted as the backbone model. As can be seen, the performance of our model generally boosts with the increasing masking rate, followed by a subsequent decline. This is reasonable, as a higher masking rate can mask more caption-related image patches, thereby enhancing the quality of pseudo-triples data; while an overly high masking rate may result in the loss of image information, making it difficult for the model to leverage visual information for reasoning. 

\begin{figure*}[h]
\small
  \centering
  \begin{minipage}{0.9\linewidth}
    \includegraphics[width=\linewidth,height=0.5\linewidth]{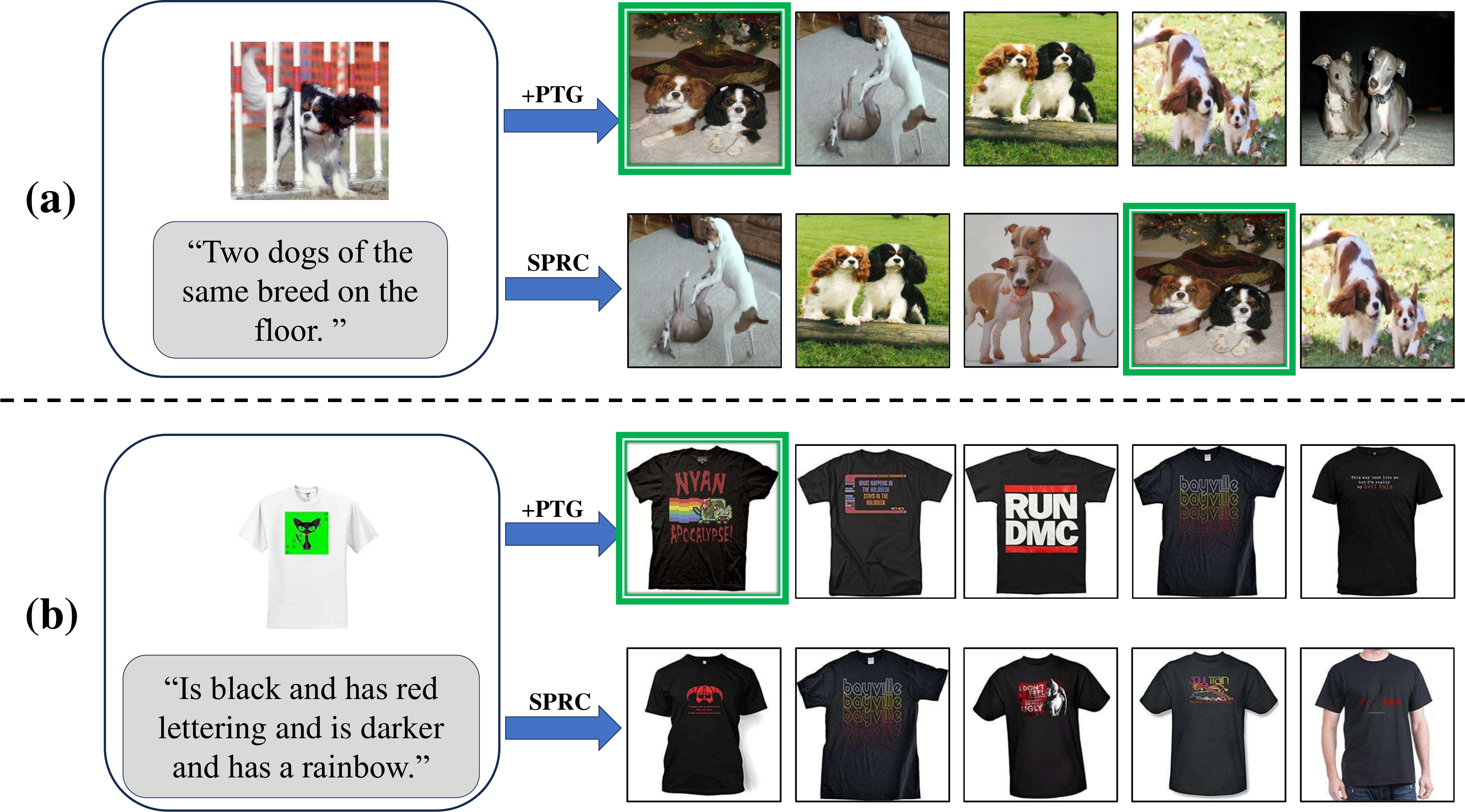}
  \end{minipage}%
  \caption{Illustration of CIR results by SPRC with or without our scheme on CIRR and FashionIQ, respectively. The ground-truth target images are highlighted with green boxes.
}  
  \label{fig:case study}
\end{figure*}

\begin{figure}[t]
  \centering
  \small
  \begin{minipage}{0.75\linewidth}
    \includegraphics[width=\linewidth]{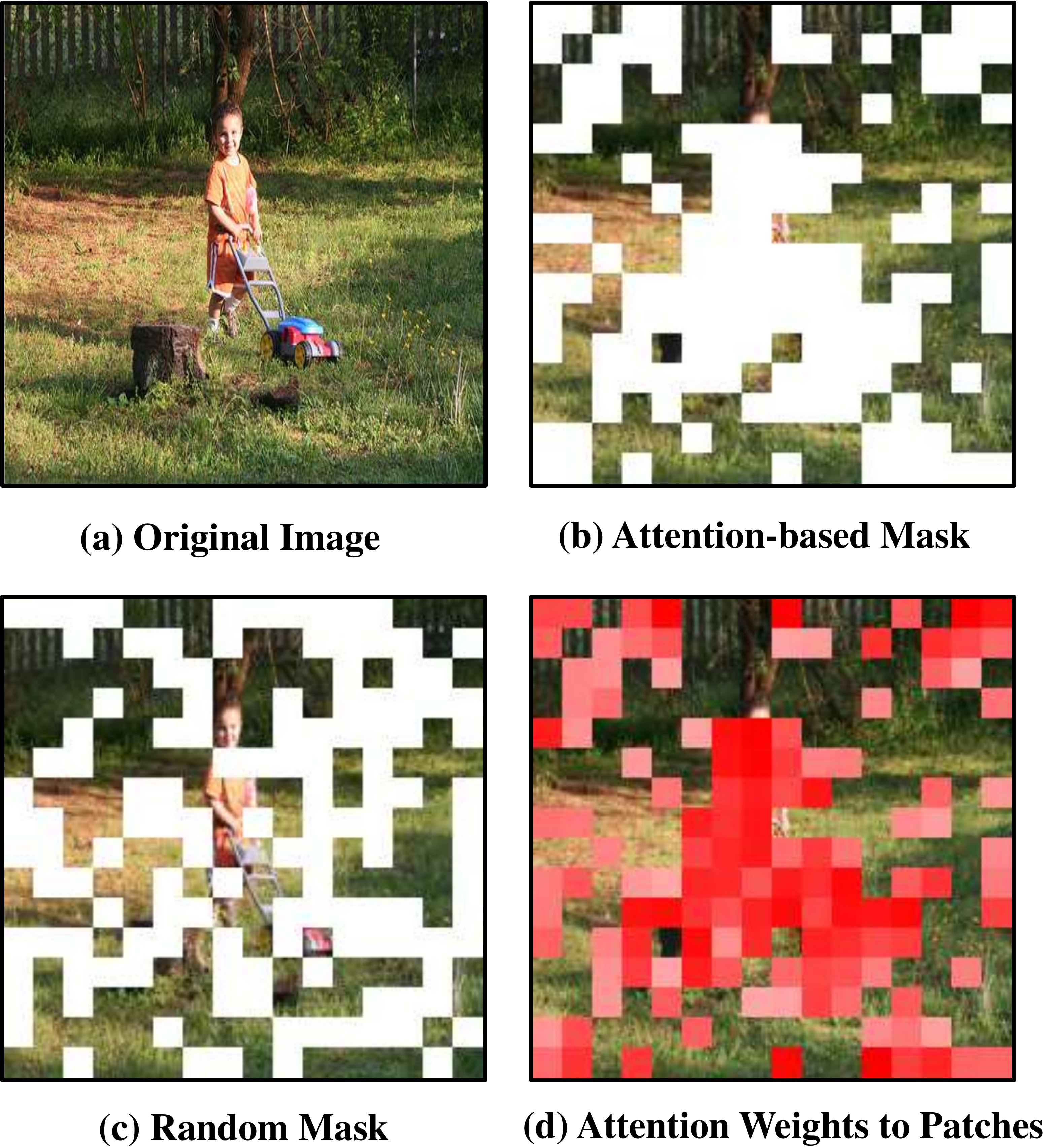}
  \end{minipage}%
\caption{Mask examples. (a) Original image (b) Attention-based mask (c) Random mask (d) Attention weight map. The darker the color, the higher the attention weight.}
\vspace{-1.5em}
  \label{fig:mask_example}
\end{figure}

\subsection{Case Study}
Figure \ref{fig:case study} illustrates two CIR results obtained by SPRC with or without our scheme in the 16-shot scenario on the fashion-domain dataset FashionIQ and the open-domain dataset CIRR. Due to the limited space, we reported the top $5$ retrieved images. 
As can be seen from the first case, shown in Figure \ref{fig:case study} (a), the user wants to change the number and position of the dog in the picture. For this straightforward modification request, both two methods correctly rank the target image in the top-$5$ places. 
Nevertheless, in Figure \ref{fig:case study} (b), the modification is more challenging, requiring not only the adjustment of the darker color but also the incorporation of a complex pattern consisting of red letters and a rainbow. In this case, our scheme enables the model to successfully rank the ground-truth target images in the top-$1$ place, while the original SPRC backbone fails to rank them within the top-$5$ places. This reconfirms the superiority of our scheme that integrates the pseudo triplet-based pretraining and challenging samples-based fine-tuning in handling challenging modifications for FS-CIR. 
\subsection{Masking Effect Visualization}

\begin{figure}[h]
  \centering
  
  \begin{minipage}{1\linewidth}
    \includegraphics[width=\linewidth]{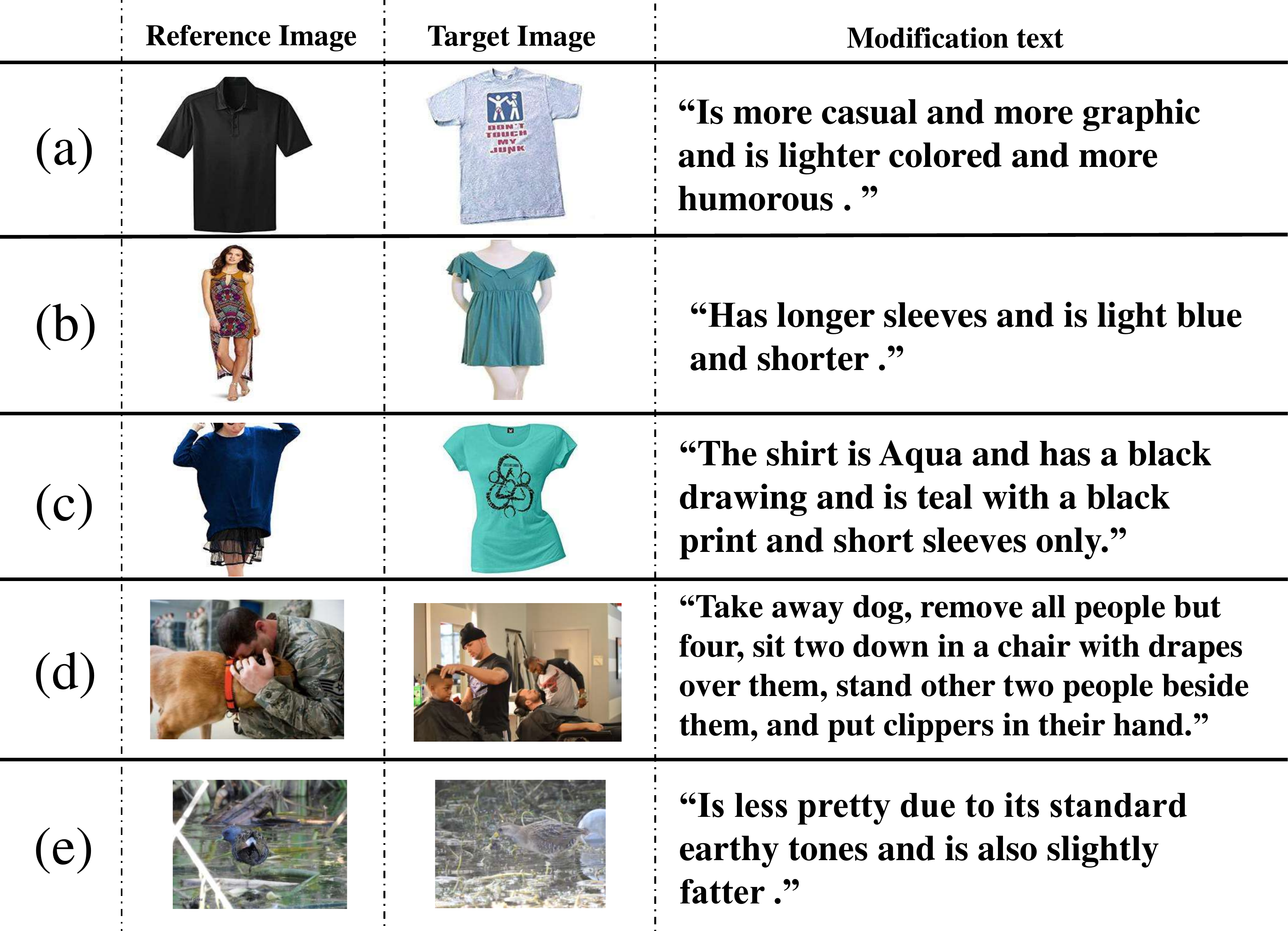}
  \end{minipage}%
  \caption{Illustration of Challenging Samples. (a)-(c) come from the three respective subsets (shirt, dress, toptee) of FashionIQ, (d) comes from CIRR and (e) comes from B2W.}
  \label{fig:selected samples}
\end{figure}

To gain a better understanding of our proposed attention-based masking strategy, we show the effect comparison between our proposed attention-based masking strategy and the random masking strategy in Figure~\ref{fig:mask_example}, where we set the masking rate to 50\% as an example. 
Specifically, Figures~\ref{fig:mask_example}(a-c) show the original image, the image masked by our proposed attention-based masking strategy, and the image masked by random masking strategy, respectively. Meanwhile, we also show the attention weights over different patches of the image (Figure~\ref{fig:mask_example}(d)), where deeper red indicates higher attention weights in the attention weight map. As can be seen, our attention-based strategy masks the main subjects of the image more precisely, namely the kid and the lawn mower, than the random masking strategy. From Figure~\ref{fig:mask_example}(d), we can also observe that the main subjects of the image, \textit{i.e.}, the kid and the lawn mower,  do have relatively high attention weights. It is reasonable to obscure the main subject as much as possible, as we aim to make the information in the masked image and the image caption complement to each other. In contrast, the random mask covers more irrelevant details, such as the yard ground, and leaves the main subject masked incompletely.  
This example illustrates the effectiveness of our proposed attention-based masking strategy. 

\subsection{Challenging Samples Visualization}
To gain deeper insights into our scheme, we illustrate several challenging samples identified by our pseudo modification text-based sample challenging score estimation method in Figure~\ref{fig:selected samples}. As can be seen, the reference and target images in all these challenging samples differ significantly, implying that these samples involve complex modifications and are difficult for the model to handle. The modifications typically involve multiple specific changes, such as style, pattern, and color in Figure~\ref{fig:selected samples}(a), sleeve length, color, and dress length in Figure~\ref{fig:selected samples}(b), as well as toptee styles and designs in Figure~\ref{fig:selected samples}(c). Moreover, in Figure~\ref{fig:selected samples}(d), the sample even involves modifications regarding the number of subjects such as ``remove all people but \textit{four}, sit \textit{two} down in a chair''. 
For the dataset B2W, in Figure~\ref{fig:selected samples}(e), we can see that when modifications involve abstract attributes of birds, such as aesthetic attributes like ``pretty'', the samples are considered challenging.

\section{Conclusion}
In this work, we propose a novel pseudo triplet-guided few-shot CIR scheme consisting of two learning stages: pseudo triplet-based CIR pretraining and challenging triplet-based CIR fine-tuning. In the first stage, we utilize an attention-based masked training strategy to construct pseudo triplets from pure image data for pretraining the model. In the second stage, a pseudo modification text-based sample challenging score estimation method is designed to select the challenging samples for fine-tuning the model pretrained in the first stage. 
It is worth noting that our scheme is plug-and-play and compatible with various CIR models. 
We conduct extensive experiments on three public datasets using two cutting-edge backbone models. Experiment results demonstrate the effectiveness of our scheme, suggesting the benefit of actively selecting discriminative samples for annotation and fine-tuning the model.


\bibliography{hbh123}
\vspace{-2em}
\begin{IEEEbiography}
[{\includegraphics[width=1in,height=1.25in,clip,keepaspectratio]{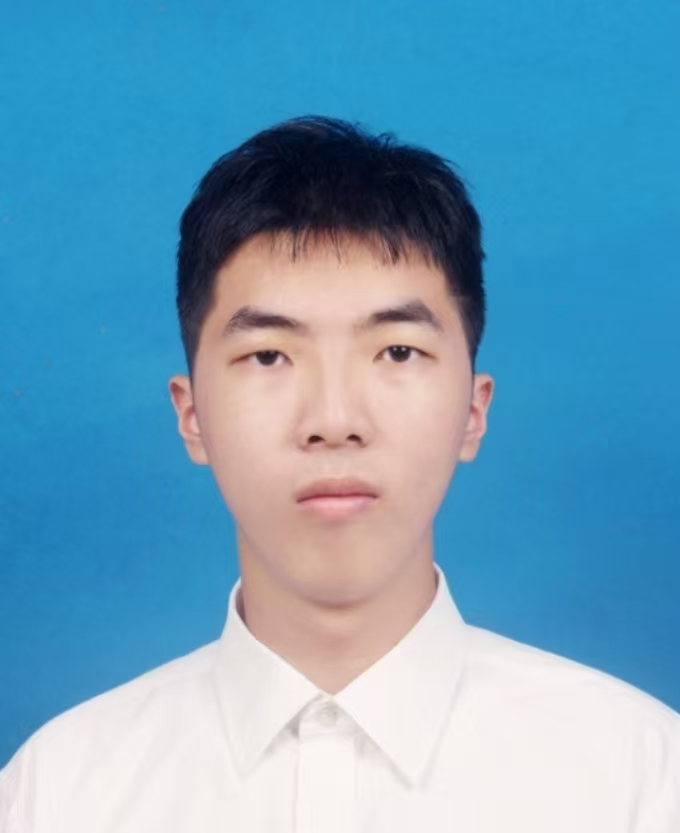}}]
{Bohan Hou} is currently pursuing the B.E. degree with the Department of Computer Science and Technology, Shandong University. He is a student of Taishan College (honorary degree) of Shandong University. His research interests include multimedia computing and information retrieval.
\end{IEEEbiography}
\vspace{-2.5em}
\begin{IEEEbiography}
[{\includegraphics[width=1in,height=1.25in,clip,keepaspectratio]{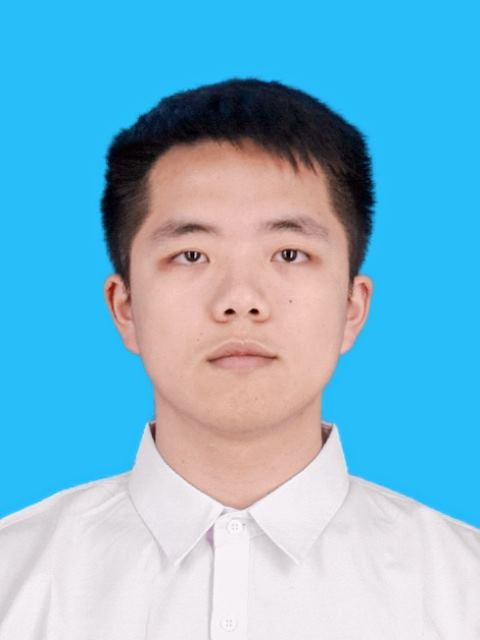}}]
{Haoqiang Lin} received the B.E. degree from the School of Computer Science and Technology, Shandong University, in 2023. He is currently pursuing the master's degree with the Department of Computer Science and Technology, Shandong University. His research interests include multimedia computing and information retrieval. He has published a paper in ACM SIGIR.
\end{IEEEbiography}
\vspace{-2.5em}
\begin{IEEEbiography}
[{\includegraphics[width=1in,height=1.25in,clip,keepaspectratio]{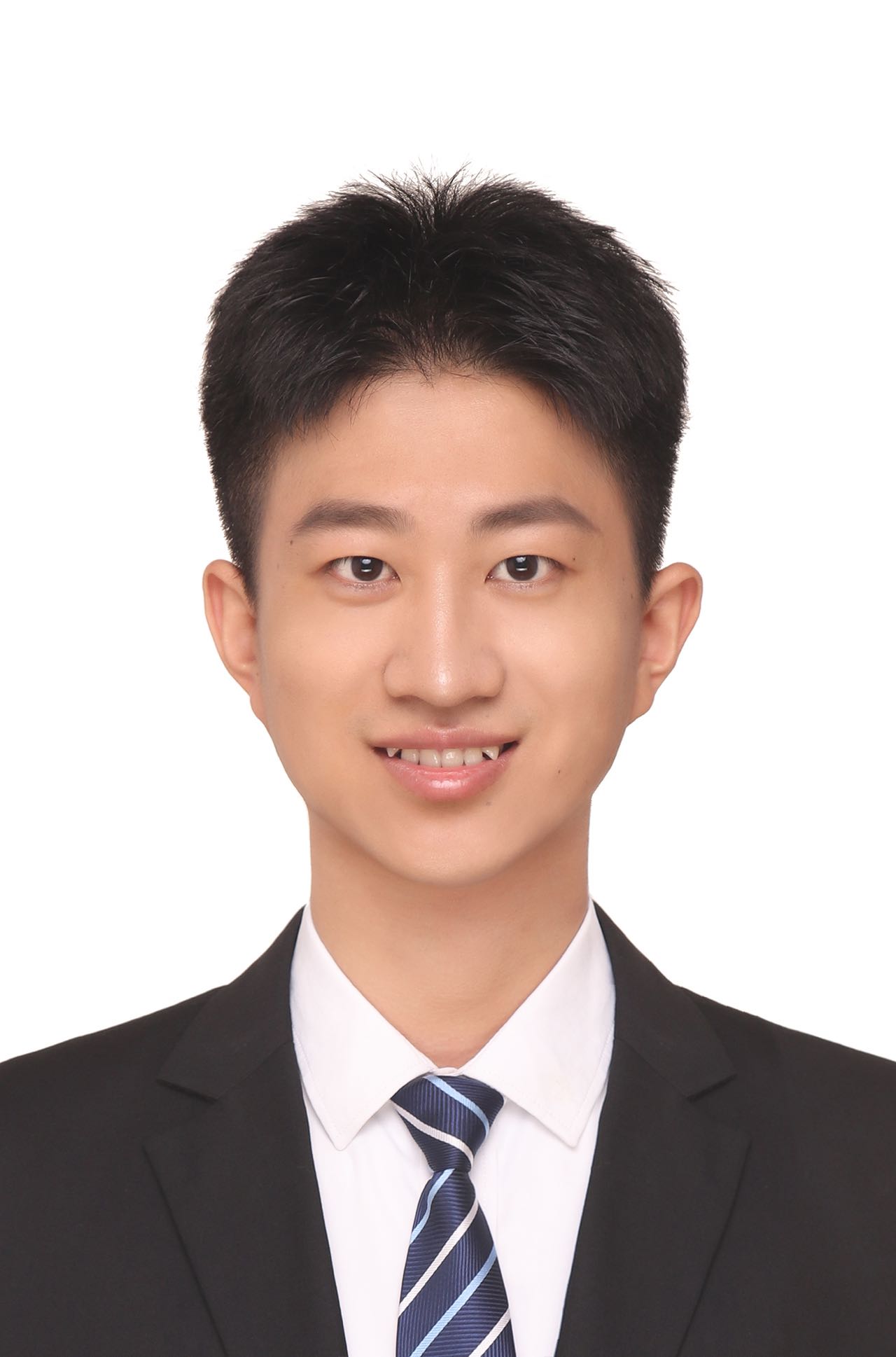}}]
{Haokun Wen} received the B.E. degree from the Ocean University of China, in 2019, and the master's degree from the School of Computer Science and Technology, Shandong University, in 2022. He is currently pursuing the Ph.D. degree with the Department of Computer Science and Technology, Harbin Institute of Technology (Shenzhen). He is also with the Department of Data Science, City University of Hong Kong, Hong Kong, China. His research interests include multimedia computing and information retrieval. He has published several papers in top venues, such as IEEE TPAMI, IEEE TIP, and ACM SIGIR.
\end{IEEEbiography}
\vspace{-2.5em}
\begin{IEEEbiography}[{\includegraphics[width=1in,height=1.25in,clip,keepaspectratio]{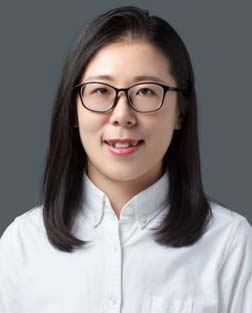}}] {Meng Liu} received the Ph.D. degree from Shandong University, China, in 2019. She is currently a Professor with the School of Computer Science and Technology, Shandong Jianzhu University. Her research interests include multimedia computing and information retrieval. Various parts of her work have been published in top conferences and journals, such as SIGIR, ACM MM, and IEEE TPAMI. She has served as a Reviewer for various conferences and journals, such as ACM MM, CVPR, and IEEE TIP.
\end{IEEEbiography}
\vspace{-2.5em}
\begin{IEEEbiography}
[{\includegraphics[width=1in,height=1.25in,clip,keepaspectratio]{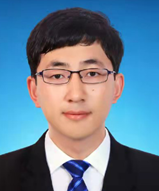}}]
{Mingzhu Xu} received the B.S., M.Sc., and Ph.D. degrees from the Harbin Institute of Technology (HIT), Harbin, China, in 2013, 2015, and 2021, respectively. He is currently an Assistant Professor with the School of Software, Shandong University, Jinan, China. His research interests include computer vision, multimedia computing, and information retrieval. Dr. Xu is also an Invited Reviewer for prestigious journals, including IEEE TMM, IEEE TCSVT, IEEE TKDE, IEEE TITS, Information Science, ACM MM, NeurlPS, ICML.
\end{IEEEbiography}
\vspace{-2.5em}
\begin{IEEEbiography}[{\includegraphics[width=1in,height=1.25in,clip,keepaspectratio]{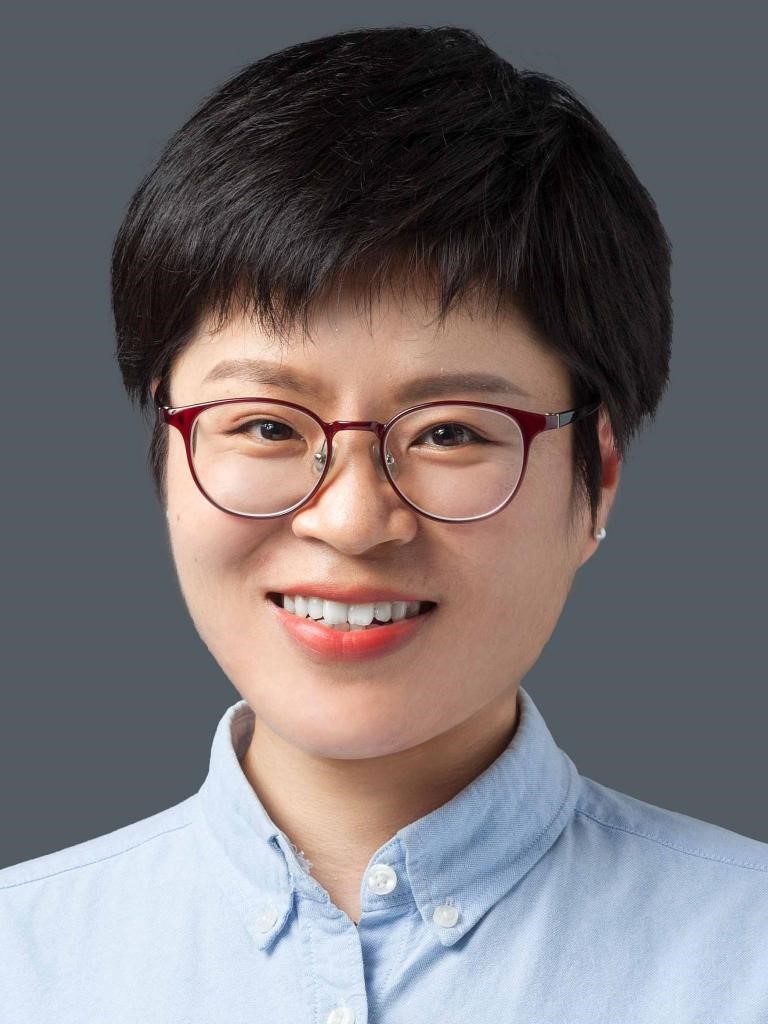}}]{Xuemeng Song}(Senior Member, IEEE)
	received the B.E. degree from the University of Science and Technology of China, in 2012, and the Ph.D. degree from the School of Computing, National University of Singapore, in 2016. She is currently an Associate Professor with Shandong University, China. She has published several papers in the top venues, such as ACM SIGIR, MM, and TOIS. Her research interests include information retrieval and social network analysis. She has served as a reviewer for many top conferences and journals. She is also an AE of IET Image Processing. 
\end{IEEEbiography}
\end{document}